\documentclass{article}
\usepackage{ijcai20}
\usepackage{amsmath,amssymb,amsfonts}
\usepackage{times}
\usepackage{soul}
\usepackage{url}
\usepackage[hidelinks]{hyperref}
\usepackage[utf8]{inputenc}
\usepackage[small]{caption}
\usepackage{graphicx}
\usepackage{amsthm}
\usepackage{booktabs}
\usepackage{algorithm}
\usepackage{algorithmic}
\usepackage{amsfonts} 
\usepackage{amsmath, bm}
\usepackage{multirow}
\usepackage{bbm}
\usepackage{color}
\usepackage{multirow}
\usepackage{subcaption}
\newtheorem{definition}{Definition}
\newtheorem{theorem}{Theorem}

\begin{document}

\title{Removing Disparate Impact of Differentially Private Stochastic Gradient Descent on Model Accuracy}

\author{
Depeng Xu
\and
Wei Du\And
Xintao Wu
\affiliations
University of Arkansas
\emails
\{depengxu,wd005,xintaowu\}@uark.edu
}

\maketitle

\begin{abstract}
When we enforce differential privacy in machine learning, the utility-privacy trade-off is different w.r.t. each group. Gradient clipping and random noise addition disproportionately affect underrepresented and complex classes and subgroups, which results in inequality in utility loss. In this work, we analyze the inequality in utility loss by differential privacy and propose a modified differentially private stochastic gradient descent (DPSGD), called DPSGD-F, to remove the potential disparate impact of differential privacy on the protected group. DPSGD-F adjusts the contribution of samples in a group depending on the group clipping bias such that differential privacy has no disparate impact on group utility. Our experimental evaluation shows how group sample size and group clipping bias affect the impact of differential privacy in DPSGD, and how adaptive clipping for each group helps to mitigate the disparate impact caused by differential privacy in DPSGD-F.

\end{abstract}

\section{Introduction}
Most researches on fairness-aware machine learning study whether  the predictive decision made by machine learning model is discriminatory against the protected group \cite{Kamishima2011FairnessAware,Zafar2017Fairness,Kamiran2010Discrimination,Hardt2016Equality,Zhang2018Mitigating,Madras2018Learning}. For example, demographic parity requires that a prediction is independent of the protected attribute.
Equality of odds \cite{Hardt2016Equality} requires that a prediction is independent of the protected attribute conditional on the original outcome.
These fairness notions focus on achieving non-discrimination within one single model.
In addition to the within-model fairness, cross-model fairness also arises in differential privacy preserving machine learning models when we compare the accuracy loss incurred by private model between the majority group and the protected group. Recently, research in \cite{Bagdasaryan2019}
shows that the reduction in accuracy incurred by deep private models disproportionately impacts underrepresented subgroups. The unfairness in this cross-model scenario is that the reduction in accuracy due to privacy protection is  discriminatory against the protected group. 

In this paper, we study   the inequality in utility loss due to differential privacy w.r.t. groups, which compares the change in prediction accuracy w.r.t. each group between the private model and the non-private model.
Differential privacy guarantees the query results or the released model cannot be exploited by attackers to derive whether one particular record is present or absent in the underlying dataset \cite{Dwork2006}. 
When we enforce   differential privacy onto a regular non-private model, the model trades some utility off for privacy. 
On one hand, with the impact of differential privacy, the within-model unfairness in the private model may be different from the one in the non-private model \cite{JagielskiKMORSU19,WWWXuYW19,ding2020differentially,CummingsGKM19}. 
On the other hand,  differential privacy may introduce additional discriminative effect towards the protected group when we compare the private model with the non-private model.
The utility  loss between the private and non-private models w.r.t. each group, such as reduction in group accuracy, may be uneven. 
The intention of differential privacy should not be to introduce more accuracy loss on the protected  group regardless of the level of within-model unfairness in the non-private model.

There are several empirical studies on  the relationship between the utility loss due to differential privacy and groups with different represented sample sizes.
Research in \cite{Bagdasaryan2019}
shows that the accuracy of private models tends to decrease more on classes that already have lower accuracy in the
original, non-private model.
In their case, the direction of inequality in utility loss due to differential privacy is the same as  the existing within-model discrimination against the underrepresented group in the non-private model, i.e. ``the poor become poorer".
Research in \cite{Du2019} shows the similar observation that the contribution of rare training
examples is hidden by random noise in  differentially private stochastic gradient descent, and that random noise slows
down the convergence of the learning algorithm.  
Research in \cite{Jaiswal2019} shows different observations when they analyze
if the performance on emotion recognition is affected in an
imbalanced way for the models trained to enhance privacy. They find that while the performance is affected differently for the subgroups, the effect is not consistent across
multiple setups and datasets. In their case, there is no consistent direction of  inequality in utility loss by differential privacy  against the underrepresented group.
Hence, the impact of differential privacy on group accuracy is more complicated than the  observation in \cite{Bagdasaryan2019} (see detailed discussions in Section \ref{sec:comments}).
It needs to be cautionary to conclude   that differential privacy introduces more utility loss on the underrepresented group.
The bottom line is that the objective of differential privacy is to protect individual's privacy instead of introducing unfairness in the form of inequality in utility loss w.r.t. groups. 
Though the privacy metric increases when a model is adversarially trained to enhance privacy, we need to ensure that
the performance of the model on that dataset does not harm
one subgroup more than the other.

In this work, we conduct theoretical analysis of the inequality in utility loss by differential privacy and propose a new differentially private mechanism to remove it. We use ``cost of privacy" to refer to the utility  loss between the private and non-private models as the result of the utility-privacy trade-off.
We study the cost of privacy  w.r.t. each group in comparison with the whole population and explain how group sample size is related to the privacy impact on group accuracy along with other factors (Section \ref{sec:exp}). 
The difference in group sample sizes leads to the difference in average group gradient norms, which results in different group clipping biases under the uniform clipping bound.
It costs less utility trade-off to achieve the same level of differential privacy for the group with larger group sample size and/or smaller group clipping bias. In other words, the group with smaller group sample size and/or larger group clipping bias incurs more utility loss when the algorithm achieves the sample level of differential privacy w.r.t. each group.
Furthermore, we propose a modified differentially private stochastic gradient descent (DPSGD) algorithm, called DPSGD-F, to remove the potential inequality in utility loss among groups (Section \ref{sec:alg}). 
DPSGD-F adjusts the contribution of samples in a group depending on the group clipping bias. For the group with smaller cost of privacy, their contribution is decreased and the achieved privacy w.r.t. their group is stronger; and vise versa. As a result, the final utility loss is the same for each group, i.e. differential privacy has no disparate impact on group utility in DPSGD-F.
Our experimental evaluation shows the effectiveness of our removal algorithm on  achieving equal utility loss with satisfactory utility (Section \ref{sec:eval}). 

Our contributions are as follows:
\begin{itemize}
    \item We provide theoretical analysis on the group level cost of privacy and  show the source of disparate impact of differential privacy on each group in the original DPSGD.
    \item We propose a modified DPSGD algorithm, called DPSGD-F,  to achieve differential privacy with equal utility loss w.r.t. each group. It uses adaptive  clipping to adjust the sample contribution of each group, so the privacy level w.r.t. each group is calibrated based on their cost of privacy.  As a result, the final group utility loss is the same for each group in DPSGD-F.
    \item In our experimental evaluation, we show  how group sample size and group clipping bias affect the impact of differential privacy in DPSGD, and how adaptive  clipping for each group helps to mitigate the disparate impact caused by differential privacy in DPSGD-F.
\end{itemize}

\section{Related Works}
\subsection{Differential Privacy}

Existing literature in differentially private machine learning targets both convex and nonconvex optimization algorithms and can be divided into three main classes, input perturbation, output perturbation, and inner perturbation.  
Input perturbation approaches \cite{DBLP:conf/focs/DuchiJW13} add noise to the input data based on local differential privacy model. 
Output perturbation approaches \cite{DBLP:conf/nips/BassilyTT18} add noise to the model after the training procedure finishes, i.e., without modifying the training algorithm.  Inner perturbation approaches modify the learning algorithm such that the noise is injected during learning. For example, research in \cite{DBLP:journals/jmlr/ChaudhuriMS11} modifies the objective of the training procedure and  \cite{AbadiCGMMT016} add noise to the gradient output of each step of the training without modifying the objective.

Limiting users to small contributions keeps noise level at the cost of introducing bias. 
Research in \cite{AminKMV19} characterizes the trade-off between bias and variance, and shows that (1) a proper bound can be found depending on properties of the dataset and (2) a concrete cost of privacy cannot be avoided simpling by collecting more data. 
Several works study how to adaptively bound the contributions of users and clip the model parameters to improve learning accuracy and robustness. 
Research in \cite{AdaCliP} uses coordinate-wise adaptive clipping of the gradient to achieve the same privacy guarantee with much less added noise. In federated learning setting, the proposed approach \cite{Thakkar2019} adaptively sets the clipping norm applied to each user's update, based on a differentially private estimate of a target quantile of the distribution of unclipped norms remove the need for such extensive parameter tuning.
Other than adaptive clipping, research in \cite{PhanWHD17} adaptively injects noise into features based on the contribution of each to the output so that  the utility of deep neural networks under
differential privacy is improved; research in \cite{LeeK18}  adaptively allocates per-iteration privacy budget 
to achieve zCDP on gradient descent.

\subsection{Fairness-aware Machine Learning}

In the literature, many methods have been proposed to modify the training data for mitigating biases and achieving fairness. These methods include: Massaging \cite{kamiran2009}, Reweighting  \cite{Calders2009Building}, Sampling  \cite{Kamiran2012Data}, Disparate Impact Removal  \cite{feldman2015}, Causation-based Removal  \cite{Zhang2017Causal} and Fair Representation Learning \cite{Edwards2015Censoring,xie2017,Madras2018Learning,Zhang2018Mitigating}.
Some researches propose to mitigate discriminative bias in model predictions by adjusting the learning process \cite{Zafar2017Fairness} or changing the predicted labels \cite{Hardt2016Equality}.
Recent studies \cite{Zhang2018Mitigating,Madras2018Learning} also use adversarial learning  to achieve fairness in classification and representation learning. 

Reweighting or sampling changes the importance of training samples according to
an estimated probability that they belong to the protected group so that more importance is placed on sensitive ones \cite{Calders2009Building,DBLP:conf/innovations/DworkHPRZ12,Kamiran2012Data}.
Adaptive sensitive reweighting uses an iterative reweighting process  to recognize sources of bias and
diminish their impact without affecting features or labels \cite{KrasanakisXPK18}.
Research in \cite{DBLP:conf/icml/KearnsNRW18} uses   agnostic learning to achieve good accuracy and fairness on all subgroups. However, it requires a large number of iterations, thus incurring a very high privacy loss. Other approaches to balance accuracy across classes include oversampling, adversarial training  with a loss function that overweights the underrepresented group, cost-sensitive learning,
and resampling. These techniques cannot be directly combined with DPSGD because the sensitivity bounds enforced by DPSGD are not valid for oversampled or overweighted inputs, i.e. the information used to find optimal balancing strategy is highly sensitive with unbounded sensitivity. 

\subsection{Differential Privacy and Fairness}
Recent works study the connection between achieving  privacy protection and fairness. Research in \cite{DBLP:conf/innovations/DworkHPRZ12} proposed a notion of fairness that is a generalization of differential privacy. Research in  \cite{DBLP:journals/datamine/HajianDMPG15} developed a pattern sanitization method that achieves $k$-anonymity and fairness. Most recently, the position paper \cite{DBLP:conf/fat/EkstrandJM18} argued for integrating recent research on fairness and non-discrimination to socio-technical systems that provide privacy protection.
Later on, several works studied how to achieve within-model fairness (demographic parity \cite{WWWXuYW19,ding2020differentially}, equality of odds \cite{JagielskiKMORSU19}, equality of opportunity \cite{CummingsGKM19}) in addition to enforcing differential privacy in the private model. 
Our work in this paper studies how to prevent disparate impact of the private model on model accuracy across different groups. 

\section{Preliminary}

Let $D$ be a dataset with $n$ tuples $x_1,x_2,\cdots,x_n$, where each tuple $x_i$ includes the information of a user $i$ on $d$ unprotected attributes $A_1, A_2, \cdots, A_d$, the protected attribute $S$, and the decision $Y$.  Let $D^k$ denote a subset of $D$ with the set of tuples with $S=k$. 
Given a set of examples $D$, the non-private model outputs a classifier $ \eta({a};{w})$
with parameter ${w}$ which minimizes the loss function $\mathcal{L}_D({w})=\tfrac{1}{n}\sum_{i=1}^n L_i({w})$.
The optimal model parameter ${{w}}^*$ is defined as: ${{w}}^* = \arg\min\limits_{w} \tfrac{1}{n}\sum_{i=1}^n L_i( {w})$.
A differentially private algorithm outputs a classifier $ \tilde{\eta}({a};\tilde{w})$ by selecting $\tilde{{w}}$  in a manner that satisfies differential privacy while keeping it close to the actual optimal  ${{w}}^*$. 

\subsection{Differential Privacy}
Differential privacy guarantees   output of a query $q$ be insensitive to the presence or absence of   one   record in a dataset. 

\begin{definition}
    \textbf{Differential privacy} \cite{Dwork2006}. A randomized mechanism $\mathcal{M}: \mathcal{D}\rightarrow \mathcal{R}$ with domain $\mathcal{D}$ and range $\mathcal{R}$ is $(\epsilon,\delta)$-differentially private if, for any pair of datasets $D,D' \in \mathcal{D}$ that differ in exactly one record, and for any subseet of outputs $\mathcal{O}\in \mathcal{R}$, we have 
	\begin{equation}\Pr(\mathcal{M}(D)\in \mathcal{O}) \leq \exp(\epsilon)\cdot \Pr(\mathcal{M}(D')\in \mathcal{O}) + \delta.\nonumber\end{equation}
\end{definition}

The parameter $\epsilon$ denotes the privacy budget, which controls the amount by which the distributions induced by $D$ and $D'$ may differ. The parameter $\delta$ is a broken probability. Smaller values of $\epsilon$ and $\delta$ indicate  stronger privacy guarantee.  
\begin{definition}
	\textbf{Global sensitivity} \cite{Dwork2006}. Given a query $q: \mathcal{D} \rightarrow \mathbb{R}$, the global sensitivity $\Delta_f$ is defined as $\Delta_f=\max_{D,D'} |q(D)-q(D')|$.
\end{definition}
The global sensitivity measures the maximum possible change in $q(D)$ when one record in the dataset changes.
The Gaussian mechanism with parameter $\sigma$ adds Gaussian noise $N(0,\sigma^2)$ to each component of the model output. 

\begin{definition}
	\textbf{Gaussian mechanism} \cite{Dwork2006}. Let $\epsilon  \in [0,1]$  be arbitrary. For $c^2> 2\log(1.25/\delta)$, the Gaussian mechanism with parameter $\sigma>c\Delta_f/\epsilon$ satisfies $(\epsilon,\delta)$-differential privacy.
\end{definition}

\subsection{Differentially Private Stochastic Gradient Descent} 
The procedure of deep learning model training is to minimize the output of a loss function through numerous stochastic gradient descent (SGD) steps. \cite{AbadiCGMMT016}  proposed a differentially private SGD algorithm (DPSGD).
DPSGD uses a clipping bound on $l_2$ norm of individual updates, aggregates the clipped updates, and then adds Gaussian noise to the aggregate.  This ensures that the iterates do not overfit to any individual user's update. 

The privacy leakage of DPSGD is measured by $(\epsilon, \delta)$, i.e., computing a bound for the privacy loss $\epsilon$ that holds with certain probability $\delta$. Each iteration $t$ of DPSGD can be considered as a privacy mechanism $\mathcal{M}_t$ that has the same pattern in terms of sensitive data access. \cite{AbadiCGMMT016} further proposed a moment accounting mechanism which calculates the aggregate privacy bound when performing SGD for multiple steps. The moments accountant computes tighter bounds for the privacy loss compared to the standard composition theorems. 
The moments accountant is tailored to the Gaussian mechanism and employs the log moment of each $\mathcal{M}_t$ to derive the bound of the total privacy loss. 
The log moment of privacy loss follows linear composability.

\begin{theorem}
\label{thm:comp}
\textbf{Composability of moments} \cite{AbadiCGMMT016}. For a given mechanism $\mathcal{M}$,  the $\lambda^\text{th}$ moment $\alpha_\mathcal{M}(\lambda)\triangleq \max\limits_{aux,D,D'}  \alpha_\mathcal{M}(\lambda;aux,D,D')$, where the maximum is taken over all possible auxiliary input $aux$ and all neighboring datasets $D,D'$. Suppose that a mechanism $\mathcal{M}$ consists of a sequence of adaptive mechanisms $\mathcal{M}_1,\ldots,\mathcal{M}_p$, where $\mathcal{M}_i: \prod_{j=1}^{i-1}\mathcal{R}_j\times \mathcal{D}\rightarrow \mathcal{R}_i$. Then, for any $\lambda$
\begin{equation}
\nonumber
    \alpha_\mathcal{M}(\lambda) \leq \sum\limits_{i=1}^p \alpha_{\mathcal{M}_i}(\lambda).
\end{equation}
\end{theorem}

\begin{algorithm}[tb]
	\begin{algorithmic}[1]
        \FOR {$t\in [T]$}
            \STATE Randomly sample a batch of samples $B_t$ with $|B_t|=b$ from $D$
            \FOR {each sample $x_i\in B_t$}
                \STATE $g_i=\triangledown {L}_i({w}_t)$
                \label{line:g}
            \ENDFOR
            \FOR {each sample $x_i\in B_t$}
                \STATE $\bar{g}_i=g_i\times \min\left(1,\frac{C}{|g_i|}\right)$
                \label{line:bar_g}
            \ENDFOR
            \STATE $\tilde{G}_B=\frac{1}{b}\left(\sum_i \bar{g}_i+N(0,\sigma^2C^2\mathbf I)\right)$
            \label{line:tilde_g}
            \STATE $\tilde{w}_{t+1}=\tilde{w}_{t}-r \tilde{G}_B$
        \ENDFOR
		\STATE Return ${\tilde{w}_T}$ and accumulated privacy cost $(\epsilon,\delta)$
	\end{algorithmic}	
	\caption{DPSGD (Dataset $D$, loss function $ \mathcal{L}_D({w})$, learning rate $r$, batch size $b$, noise scale $\sigma$, clipping bound $C$)}
	\label{alg:dpsgd}
\end{algorithm}

To reduce noise in private training of neural networks, DPSGD \cite{AbadiCGMMT016} truncates the gradient of a neural network to control the sensitivity of the sum of gradients. This is because the sensitivity of gradients and the scale of the noise would otherwise be unbounded. To fix this,  a cap $C$ on the maximum size of a user's contribution is adopted (Line 7 in Algorithm \ref{alg:dpsgd}).  This will bias our estimated sum but also reduce  the amount of added noise, as the sensitivity of the sum is now $C$.  One  question is how to choose the truncation level for the gradient norm. If set  too high, the noise level may be so great that any utility in the result is lost. If set too low, a large amount of gradients will be forced to clip. DPSGD simply suggests using the median of observed gradients. 
\cite{AminKMV19} investigated this bias-variance trade-off and showed that the limit we should choose is the $(1-1/b\epsilon)$-quantile of the gradients themselves. It does
not matter how large or small the gradients are above
or below the cutoff, only that a fixed number of values
are clipped.

\subsection{Within-model Fairness}
Consider the classifier $\eta: {A} \rightarrow Y$ which predicts the class label $Y$ given the unprotected attributes ${A}$. Classification fairness requires that the predicted label $\eta({A})$ is unbiased with respect to the protected variable $S$. The following notions of fairness in classification was defined by \cite{Hardt2016Equality} and refined by \cite{Beutel2017Data}.

\begin{definition}
\textbf{Demographic parity}
Given a labeled dataset ${D}$ and a classifier $\eta: {A} \rightarrow Y$, the property of demographic parity is defined as 
\begin{equation}
\nonumber
   P(\eta({A})=1|S=i)=P(\eta({A})=1|S=j).
\end{equation}
\end{definition}
This means that the predicted labels are independent of the protected attribute.

\begin{definition}
\textbf{Equality of odds}
Given a labeled dataset ${D}$ and a classifier $\eta$, the property of  equality of odds is defined as
\begin{equation}
\nonumber
   {P(\eta({A})=1|Y=y,S=i) }= P(\eta({A})=1|Y=y,S=j),
\end{equation}
where $y \in \{0,1\}$.
\end{definition}
Hence, for $Y=1$,  equality of odds requires the classifier $\eta$ has equal true positive rates (TPR) between two subgroups $S=i$ and $S=j$; for $Y=0$, the classifier $\eta$ has equal false positive rates (FPR) between two subgroups. 

Equality of odds promotes that individuals who qualify for  a desirable outcome should have an equal chance of being correctly classified for this outcome. It allows for higher accuracy with respect to non-discrimination. It enforces both equal true positive rates and false positive rates in all demographics, punishing models that perform well only on the majority.

\section{Disparate Impact on Model Accuracy}
In this section, we first discuss how differentially private learning, specifically DPSGD, causes inequality in utility loss through our preliminary observations.
Then we study the cost of privacy with respect to each group in comparison with the whole population and explain how group sample size is related to the privacy impact  on group accuracy along with other factors.

\begin{table*}[]
\small
\centering
\begin{tabular}{|c|c|c|c|c|c|c|c|c|c|}
\hline
Dataset       & \multicolumn{3}{c|}{MNIST}  & \multicolumn{3}{c|}{Adult}  & \multicolumn{3}{c|}{Dutch}  \\ \hline
Group         & Total   & Class 2 & Class 8 & Total   & M       & F       & Total   & M       & F       \\ \hline
Sample size   & 54649   & 5958    & 500     & 45222   & 30527   & 14695   & 60420   & 30273   & 30147   \\ \hline
SGD           & 0.9855  & 0.9903  & 0.9292  & 0.8099  & 0.7610  & 0.9117  & 0.7879  & 0.8013  & 0.7744  \\ \hline
DPSGD         & 0.8774  & 0.9196  & 0.2485  & 0.7507  & 0.6870  & 0.8836  & 0.6878  & 0.6479  & 0.7278  \\ \hline
DPSGD vs. SGD & -0.1081 & -0.0707 & -0.6807 & -0.0592 & -0.0740 & -0.0281 & -0.1001 & -0.1534 & -0.0466 \\ \hline
\end{tabular}
	\caption{Model accuracy w.r.t. the total population, the majority group and the minority group for SGD and DPSGD on the unbalanced MNIST ($\epsilon=6.55, \delta=10^{-6}$), the original Adult ($\epsilon=3.1, \delta=10^{-6}$) and the original Dutch ($\epsilon=2.66, \delta=10^{-6}$) datasets}
	\label{tbl:original}
\end{table*}

\begin{figure*}
         \centering
         \includegraphics[width=\textwidth]{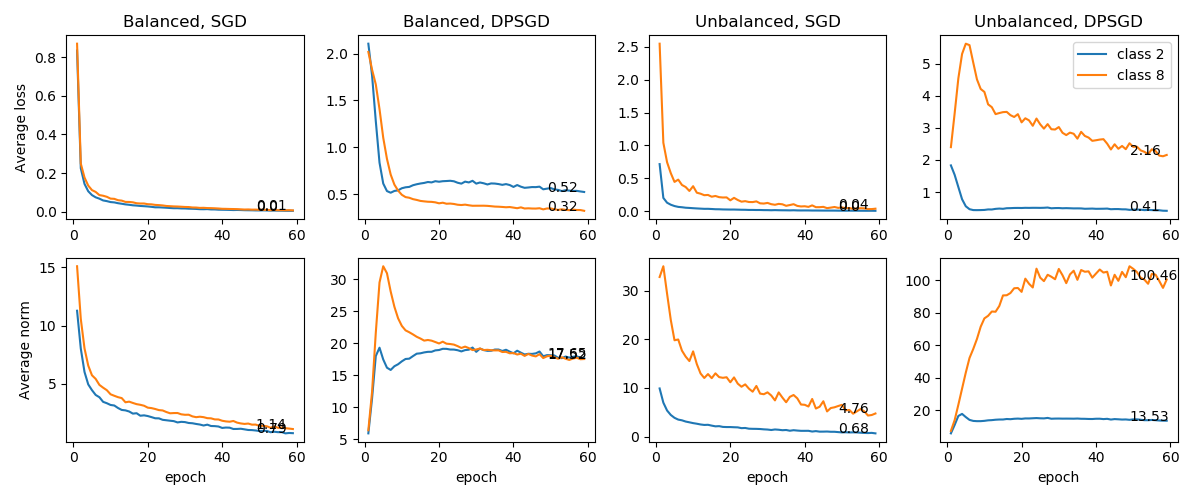}
    	\caption{The average loss  and  the average  gradient norm w.r.t. class 2 and 8 over epochs for SGD and DPSGD on the MNIST dataset (Balanced: $\epsilon=6.23, \delta=10^{-6}$, Unbalanced: $\epsilon=6.55, \delta=10^{-6}$)}
    	\label{fig:group_mnist}
\end{figure*}

\begin{figure*}
         \centering
         \includegraphics[width=\textwidth]{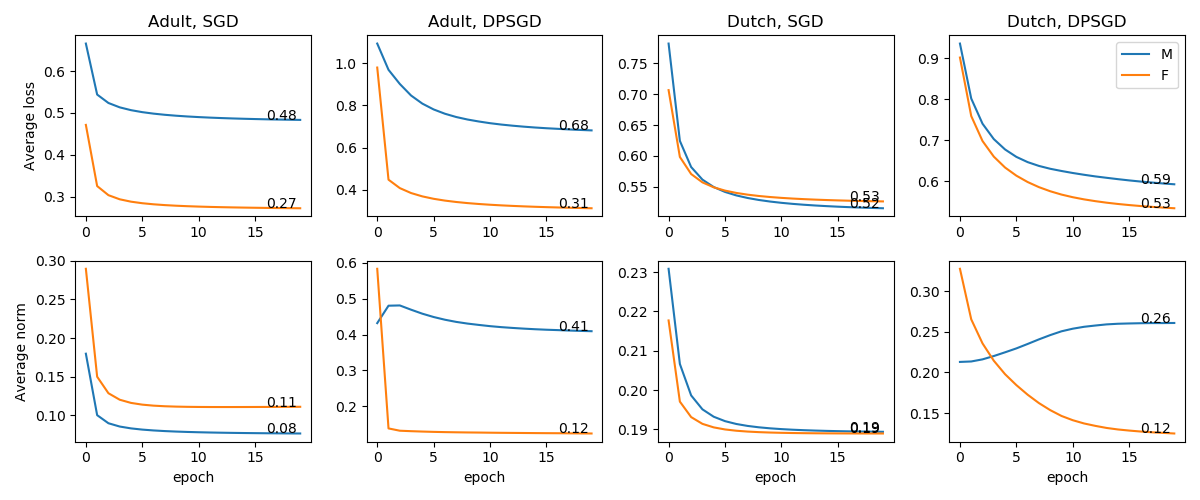}
    	\caption{The average loss  and  the average  gradient norm w.r.t. male and female groups over epochs for SGD and DPSGD on the original Adult and the original Dutch datasets (Adult: $\epsilon=3.1, \delta=10^{-6}$, Dutch: $\epsilon=2.66, \delta=10^{-6}$)}
    	\label{fig:group_adult}
\end{figure*}

\subsection{Preliminary Observations}
\label{sec:comments}

To explain why DPSGD has disparate impact on model accuracy w.r.t. each group, \cite{Bagdasaryan2019}  constructs an unbalanced MNIST dataset to study the effects of gradient clipping, noise addition, the size of the underrepresented group, batch size, length of training, and other hyperparameters.  Training on the data of the underrepresented subgroups produces larger gradients, thus clipping reduces their learning rate and the influence of their data on the model. They also show random noise addition has the biggest impact on the underrepresented inputs. 
However,  \cite{Jaiswal2019} reports inconsistent observations on whether differential privacy has negative discrimination towards the underrepresented group in terms of reduction in accuracy.
To complement their observations, we use the unbalanced MNIST dataset used in  \cite{Bagdasaryan2019} to reproduce their result, and we also use two benchmark census datasets (Adult and Dutch) in fair machine learning to study the inequality of utility loss due to differential privacy. We include the setup details in Section \ref{sec:setup}. Table \ref{tbl:original} shows the model accuracy  w.r.t. the total population, the majority group and the minority group for SGD and DPSGD on the MNIST, Adult and Dutch datasets. 

On the unbalanced MNIST dataset, the minority group (class 8) has significantly larger utility loss than the other groups in private model. DPSGD only results in $-0.0707$ decrease in accuracy on the well-represented classes  but accuracy
on the underrepresented class drops $-0.6807$, exhibiting a disparate impact on the underrepresented class. 
Figure \ref{fig:group_mnist} shows that the small sample size reduces both the convergence rate and the optimal utility of class 8 in DPSGD in comparison with the non-private SGD. 
The model is far from converging, yet
clipping and noise addition do not let it move closer to the minimum of the loss function. Furthermore, the
addition of noise, whose magnitude is similar to the update vector, prevents the clipped gradients of
the underrepresented class from sufficiently updating the relevant parts of the model. 
Training with more epochs does not reduce
this gap while exhausting the privacy budget.
Differential privacy also slows down the convergence and degrades the utility for each group. 
Hence, DPSGD introduces negative discrimination against the minority group (which  already has lower accuracy in the non-private SGD model) on the unbalanced MNIST dataset. This matches the observation in \cite{Bagdasaryan2019}.

However, on the Adult and Dutch datasets, we have  different observations from MNIST. The Adult dataset is an unbalanced dataset, where the female group is underrepresented. Even though the male group is  the majority group, it has lower accuracy in the SGD and more utility loss in DPSGD than the female group. 
The Dutch dataset is a balanced dataset, where the group sample sizes are similar for male and female. However, DPSGD introduces more negative discrimination against the male group and its direction (male group loses more accuracy due to DP) is even opposite to the direction of within-model discrimination (female group has less accuracy in SGD). 
Figure \ref{fig:group_adult} shows that the average gradient norm is much higher for the male group in DPSGD on both datasets. It is not simply against the group with smaller sample size or lower accuracy in the SGD. Hence,  differential privacy does not always introduce more accuracy loss to the minority group on the Adult and Dutch datasets. This matches the observation in \cite{Jaiswal2019}.

From the preliminary observations, we learn that the disproportionate effect from differential privacy is not guaranteed towards the underrepresented group or the group with ``poor" accuracy. 
Why does differential privacy cause inequality in utility loss w.r.t. each group? 
It may depend on more than just the represented sample size of each group: the classification model, the mechanism to achieve differential privacy, and the relative complexity of data distribution of each group subject to the model. 
One common observation among all settings is that  the group with more utility loss has larger gradients and worse convergence. 
In Figure \ref{fig:group_mnist}, the underrepresented class 8 has average gradient norm of over 100 and bad utility in DPSGD. In Figure \ref{fig:group_adult}, the male group has much larger average gradient norm than the female group in DPSGD on both Adult and Dutch datasets.
It is important to address the larger gradients  and worse convergence directly in order to mitigate the inequality in utility loss.

\subsection{Analysis on  Cost of Privacy w.r.t. Each Group}
\label{sec:exp}

In this section, 
we conduct analysis on the cost of privacy from the viewpoint of a single batch, where the utility loss is measured by the expected error of the estimated private gradient  w.r.t. each group. For ease of discussion, our analysis follows
\cite{AminKMV19} that investigates the bias-variance trade-off due to clipping in DPSGD with Laplace noise. Suppose that $B_t$ is a collection of $b$ samples, $x_1,\cdots,x_b$.  Each $x_i$ corresponds to a sample   and generates the gradient $g_i$. 
We would like to estimate the average   gradient $G_B$ from $B_t$ in a differentially private way while minimizing the objective function.  

We denote the original gradient before clipping $G_B =\frac{1}{b} \sum_{i=1}^{b} g_i$, the gradient after clipping but before adding noise $\bar{G}_B =\frac{1}{b} \sum_{i=1}^{b} \bar{g}_i$, and the gradient after clipping and adding noise $\tilde{G}_B = \frac{1}{b}(\sum_{i=1}^{b} \bar{g}_i+Lap(\frac{C}{\epsilon}))$. 
The expected error of the estimate 
$\tilde{G}_B$ consists of a variance term (due to the noise) and a bias term (due to the contribution limit): 
\begin{equation}
\begin{split}   
    \mathbb{E}|\tilde{G}_B - G_B| & \leq \mathbb{E}|\tilde{G}_B - \bar{G}_B| + |\bar{G}_B - {G}_B|\\  \nonumber
    & \leq \frac{1}{b}\frac{C}{\epsilon} + \frac{1}{b}\sum_{i=1}^{b} \max(0, |g_i|-C). 
\end{split}
\end{equation}
In the above derivation, we base the fact that the mean absolute deviation of a Laplace variable is equal to its scale parameter. We can find the optimal $C$
by noting that the bound is convex with sub-derivative ${\frac{1}{\epsilon} - |\{i : g_i > C\}|}$, thus the minimum is achieved when $C$ is equal to the $\lceil 1/\epsilon \rceil$th largest value in gradients.

The expected error is tight as we have 
\begin{equation}
    \mathbb{E}|\tilde{G}_B - G_B|\geq \frac{1}{2}\left[\frac{1}{b}\frac{C}{\epsilon} + \frac{1}{b}\sum_{i=1}^{b} max(0, |g_i|-C)\right].
    \nonumber
\end{equation}
In other words,  the limit we should choose is just the ${(1-1/b\epsilon)}$-quantile of the gradients themselves. 

For the same batch of samples, we derive the cost of privacy w.r.t. each group. Suppose the batch of samples $B_t$ are from $K$ groups and group $k$ has sample size $b^k$. We have $G_B^k= \frac{1}{b^k}\sum_{i=1}^{b^k} g_i^k$ and $G_B=\dfrac{1}{b}\sum_{k=1}^K b^k G_B^k$. 
DPSGD bounds the sensitivity of gradient by clipping each sample's gradient with a clipping bound $C$.
$\bar{G}_B^k=\frac{1}{b^k}\sum_{i=1}^{b^k} \bar{g}_i^k=\frac{1}{b^k}\sum_{i=1}^{b^k} g_i^k\times\min(1, \frac{C}{|g_i^k|})$. 
Then, DPSGD adds Laplace noise   on the sum of clipped gradients.
$\tilde{G}_B^k=\frac{1}{b^k}(b^k\bar{G}_B^k + Lap(\frac{C}{\epsilon}))$.
The expected error of the estimate 
$\tilde{G}_B^k$  also consists of a variance term (due to the noise) and a bias term (due to the contribution limit):
\begin{equation}
\begin{split}   
    \mathbb{E}|\tilde{G}_B^k - G_B^k| & \leq \mathbb{E}|\tilde{G}_B^k - \bar{G}_B^k| + |\bar{G}_B^k - {G}_B^k|\\  
    & \leq \frac{1}{b^k}\frac{C}{\epsilon} + \frac{1}{b^k}\sum_i^{b^k} \max(0, |g_i^k|-C)\\
    & = \frac{1}{b^k}\frac{C}{\epsilon} + \frac{1}{b^k}\sum_{i}^{m^k} (|g_i^k|-C),
    \label{eq:cost}
\end{split}
\end{equation}
where $m^k=|\{i:|g_i^k|>C\}|$ is the number of examples that get clipped in group $k$.
Similarly, we can get the tight bound w.r.t. each group $k$ is $\mathbb{E}|\tilde{G}_B^k - G_B^k|\geq \frac{1}{2}\left[\frac{1}{b^k}\frac{C}{\epsilon} + \frac{1}{b^k} \sum_i^{b^k} \max(0, |g_i^k|-C)\right]$.

From Equation \ref{eq:cost}, we know the utility loss of group $k$, measured by the expected error of the estimated private gradient, is bounded by two terms, the bias $\frac{1}{b^k} \sum_i^{b^k} \max(0, |g_i^k|-C)$ due to contribution limit (depending on the size of gradients and the size of clipping bound) and the variance of the noise $\frac{1}{b^k}\frac{C}{\epsilon}$  (depending on the scale of the noise). Next, we discuss their separate impacts  in DPSGD.

Given the clipping bound $C$, the bias due to clipping w.r.t. the group with  large gradients   is larger than the one w.r.t. the group with small gradients. 
Before clipping, the group with large gradients  has large contribution in the total gradient $G_B$ in SGD, but it is not the case in DPSGD.
The direction of the  total  gradient after clipping $\bar{G}_B$ is closer to the direction of the gradient of the group with small bias (small gradients) in comparison with the direction of the total gradient before clipping $G_B$. 
Due to clipping, the contribution and convergence of the group with large gradients are  reduced.

The added noise increases the variance of the model gradient, as it tries to hide the influence of a single record on the model. It  slows down the convergence rate of the model. 
Because the noise scales$\frac{C}{\epsilon}$ and the sensitivity of clipped gradients  $C$ are the same for all groups,  the noisy gradients of all groups achieve the same level of differential privacy $\epsilon$. 
The direction of the noise is random, i.e. it does not favor a particular group in expectation.

Overall in DPSGD,  the group with large gradients has larger cost of privacy, i.e. they have more utility loss to achieve $\epsilon$ level of differential privacy under the same clipping bound $C$. 

We can also consider the optimal choice of $C$ which is $(1-\frac{1}{b\epsilon})$-quantile for the whole batch. For each group, the optimal choice of $C^k$ is $(1-\frac{1}{b^k\epsilon})$-quantile for group $k$. The distance between $C$ and $C^k$ is not the same for all groups, and $C$ is closer to the choice of $C^k$ for the  group with small bias (small gradients).

Now we look back on the preliminary observations in Section \ref{sec:comments}. On  MNIST, 
the group sample size affects the convergence rate for each group. The group with large sample size (the majority group, class 2) has larger contribution in the total gradient than the group with small sample size (the minority group, class 8), and therefore it leads to a relatively faster and better convergence.
As the result, the gradients of the minority group are larger than the gradients of the majority group later on.  In their case, the small  sample size is the main cause of large gradient norm and large utility loss in class 8. On Adult and Dutch,
the average bias due to clipping for each group is different because the distributions of gradients are quite different. The average gradient norm of the male group is larger than the average gradient norm of the female group, even though the male group is not underrepresented. As the result, the male group's contribution is limited due to clipping and it has larger utility loss in DPSGD. In there case, the group sample size is not the only reason to cause difference in  the average gradient norm, and   the other factors (e.g., the relative complexity of data distribution of each group subject to the model) out-weighs sample size, so the well-represented male group has larger utility loss.

This gives us an insight on the relation between differential privacy and the inequality in utility loss w.r.t. each group. 
The direct cause of the inequality is the large cost of privacy due to large average gradient norm (which can be caused by small group sample size along with other factors). 
In DPSGD, the clipping bound is selected uniformly for each group without consideration of the difference in clipping biases. As a result, the noise addition to achieve $(\epsilon ,\delta)$-differential privacy on the learning model results in different utility-privacy trade-off for each group, where the underrepresented or the more complex group  incurs a larger utility loss. 
After all, DPSGD is designed to protect individual's privacy with nice properties without consideration of its different impact towards each group. 
In order to avoid disparate utility loss among groups, we need to modify DPSGD such that each group needs to achieve different level of privacy to counter their difference in  costs of privacy.

\section{Removing Disparate Impact}

Our objective is to build a learning algorithm that outputs a neural network  classifier $\tilde{\eta} ({a};\tilde{w})$
with parameter $\tilde{w}$ that achieves differential privacy and equality of utility loss with satisfactory utility. 
Based on our preliminary observation and analysis on cost of privacy in DPSGD, we propose a heuristic removal algorithm to achieve equal utility loss w.r.t. each group, called DPSGD-F.

\subsection{Equality of Impact of Differential Privacy}
In the within-model fairness, equality of odds results in the equality of accuracy for different groups. Note that equal accuracy does not result in equal odds. As a trade-off for privacy, differential privacy results in accuracy loss on the model. However, different groups may incur different levels of  accuracy loss. 
We use reduction in accuracy w.r.t. group $k$ to measure utility loss between the private model $\tilde{\eta}$ and the non-privacte model $\eta$, denoted by $\Delta^k$.
We define a new fairness notion called \textit{equality of privacy impact} for differentially private learning, which requires that the  utility loss due to differential privacy is the same for all groups.

\begin{definition}\textbf{Equality of privacy impact}
\label{def:eoi}
Given a labeled dataset ${D}$, a classifier $\eta$ and a differentially private classifier $\tilde{\eta}$, a differentially private mechanism  satisfies equality of privacy impact  if
\begin{equation}
    \Delta^i(\tilde{\eta}-\eta)=\Delta^j(\tilde{\eta}-\eta),
    \nonumber
\end{equation}
where  $i,j$  are any two values of the protected attribute $S$. 
\end{definition}

\subsection{Removal Algorithm}
\label{sec:alg}

\begin{algorithm}[tb]
	\begin{algorithmic}[1]
        \FOR {$t\in [T]$}
            \STATE Randomly sample a batch of samples $B_t$ with $|B_t|=b$ from $D$ 
            \FOR {each sample $x_i \in B_t$}
                \STATE $g_i=\triangledown L_i({w}_t)$
            \ENDFOR
            \FOR {each group $k \in [K]$}
            \label{line:begin}
                \STATE $m^k=\left|\left\{i:|g_i^k|>C_0\right\}\right|$
                \STATE $o^k=\left|\left\{i:|g_i^k|\leq C_0\right\}\right|$
            \ENDFOR
            \STATE $\left\{\tilde m^k,\tilde o^k\right\}_{k \in [K]}=\left\{m^k,o^k\right\}_{k \in [K]}+N(0,\sigma_1^2\mathbf I)$
            \label{line:sigma1}
            \STATE $\tilde{m} = {\sum_{k \in [K]} \tilde m^k}$
            \FOR {each group $k \in [K]$}
                \STATE $\tilde{b}^k = \tilde m^k+\tilde o^k$
                \label{line:end}
                \STATE $C^k=  {C_0}\times \left(1+\frac{\tilde m^k/\tilde b^k}{\tilde{m}/b}\right)$
                \label{line:adapt}
            \ENDFOR
            \FOR {each sample $x_i \in B_t$}
                \STATE $\bar{g}_i= g_i\times\min\left(1,\frac{C^k}{|g_i|}\right)$
            \ENDFOR
            \STATE $C=\max\limits_k C^k$
            \STATE $\tilde{G}_B=\frac{1}{b}\left(\sum_i \bar{g}_i+N(0,\sigma_2^2C^2\mathbf I)\right)$
            \label{line:sigma2}
            \STATE $\tilde{w}_{t+1}=\tilde{w}_{t}-r \tilde{G}_B$
        \ENDFOR
		\STATE Return ${\tilde{w}_T}$ and accumulated privacy cost $(\epsilon,\delta)$
		\label{line:return}
	\end{algorithmic}	
	\caption{DPSGD-F (Dataset $D$, loss function $ \mathcal{L}_D({w})$, learning rate $r$, batch size $b$, noise scales $\sigma_1, \sigma_2$, base clipping bound $C_0$)}
	\label{alg:dpsgd_f}
\end{algorithm}

We propose a heuristic approach for differentially private SGD that removes disparate impact across different groups.  The intuition of our heuristic approach is to balance the level of privacy w.r.t. each group based on their utility-privacy trade-off. Algorithm \ref{alg:dpsgd_f} shows the framework of our approach. 
Instead of uniformly clipping the gradients for all groups, we propose to do adaptive sensitive clipping where each group $k$ gets its own clipping bound $C^k$. 
For the group with larger clipping bias (due to large gradients), we choose a larger clipping bound to balance their higher cost of privacy. The large gradients may be due to group sample size or other factors.

Based on our observation and analysis in the previous section, to balance the difference in costs of privacy for each group, we need to adjust  the clipping bound $C^k$ such that the contribution of each group is proportional to the size of their average gradient (Line \ref{line:adapt} in Algorithm \ref{alg:dpsgd_f}).
Ideally, we would like to adjust the clipping bound based on the private estimate of  the average gradient norm. However,  the original gradient before clipping has unbounded sensitivity. It would not be practical to get its private estimate. We need to construct a good approximate estimate of the relative size of the average gradient w.r.t. each group and it needs to have a small sensitivity for private estimation.

In our algorithm,
we choose adaptive clipping bound $C^k$ based on the $m^k$, where $m^k=|\{i:|g_i^k|>C_0\}|$.
To avoid the influence of group sample size, we use the fraction of $\frac{m^k}{b^k}$ that represents the fraction of samples in the group with gradients larger than $C_0$.  The relative ratio of $\frac{m^k}{b^k}$ and $\frac{m}{b}$ can approximately represent the relative size of the average gradient (Line \ref{line:adapt}). 
To choose the clipping bound $C^k$ for group $k$ in a differentially private way,
we get the  private ${\tilde m^k},{\tilde b^k}$ and $\tilde m$ from the collection $\{ m^k, o^k\}_{k \in [K]}$ (Line \ref{line:begin}-\ref{line:end}).
The collection $\{m^k,o^k\}_{k \in [K]}$ has sensitivity of 1, which is much smaller than the sensitivity of the actual gradients when we estimate the relative size of the average gradient. 

After the adaptive clipping, the sensitivity of the clipped gradient of group $k$ is $ C^k={C_0}\times (1+\frac{\tilde m^k/\tilde b^k}{\tilde{m}/b}) $. The sensitivity of the clipped gradient of the total population would be $\max_k C^k$ as the worst case in the total population needs to be considered.

Note that in Algorithm \ref{alg:dpsgd_f} we have two steps of adding noise in each iteration $t$. We first use a relatively large noise scale $\sigma_1$ (small privacy budget) to get a private collection $\{\tilde m^k,\tilde o^k\}_{k \in [K]}$ (Line \ref{line:sigma1}). Then we use a relatively small noise scale $\sigma_2$ to perturb the gradients (Line \ref{line:sigma2}). The composition theorem (Theorem \ref{thm:comp}) is applied when we compute the accumulated privacy cost $(\epsilon,\delta)$ from moments accountant (Line \ref{line:return}). Because $\sigma_1>\sigma_2$, only a small fraction of privacy budget is spent on getting $C^k$.

For the total population, Algorithm \ref{alg:dpsgd_f} still satisfies $(\epsilon,\delta)$-differential privacy as it accounts for the worst clipping bound $\max\limits_k C^k$. On the group level, each group achieves different levels of privacy depending on their utility-privacy trade-off.

With our modified DPSGD algorithm, we continue our discussion in Secion \ref{sec:exp}.
In the case of \cite{Bagdasaryan2019}, the difference in gradient norms is primarily decided by group sample size. 
Consider a majority group $s^+$ and a minority group $s^-$.
In Algorithm \ref{alg:dpsgd}, each group achieves the same level of privacy, but the underrepresented group $s^-$ has higher privacy cost (utility loss).
In Algorithm \ref{alg:dpsgd_f}, we choose a higher clipping bound $C^-$ for the  underrepresented  group. Because the noise scale is $\frac{C}{\epsilon}=\frac{C^-}{\epsilon}$ and the sensitivity of clipped gradients for the underrepresented group is $C^-$,  the noisy gradient w.r.t. the underrepresented group   achieves $\epsilon$-differential privacy. The well-represented group $s^+$ has a smaller cost of privacy, so we choose a  lower clipping bound $C^+$. Because the noise scale is $\frac{C}{\epsilon}=\frac{C^-}{\epsilon}$ and the sensitivity of clipped gradients for the underrepresented group is $C^+$,  the noisy gradient w.r.t. the underrepresented group   then  achieves $(\frac{C^+}{C^-}\epsilon)$-differential privacy. 
Two groups have  different clipping bounds $C^+,C^-$ and the same noise addition based on $C=\max(C^+,C^-)$ (same $\epsilon$ but different relative scales w.r.t. their group sensitivities). 
Hence, when we enforce the same level of utility loss for groups with different sample sizes, the well-represented group achieves stronger privacy (smaller than $\epsilon$) than the underrepresented group.
In the case of  Adult/Dutch, the male group has larger gradients regardless of the sample size.
The group with smaller gradients based on model and data distribution has smaller cost of privacy.
Algorithm \ref{alg:dpsgd_f} can adjust the clipping bound for each group. As a result, the group with smaller gradients 
achieves stronger level of privacy. Eventually, they can have similar clipping bias to  the ones in  Algorithm \ref{alg:dpsgd}.

\subsection{Baseline}
\label{sec:naive}

There is no previous work on how to achieve equal utility loss in DPSGD. 
For experimental evaluation, we also present a na\"ive baseline algorithm based on reweighting (shown as Algorithm \ref{alg:naive}) in this section, since reweighting is a common way to mitigating biases in machine learning. 
The na\"ive algorithm considers group sample size as the main cause of disproportional impact in DPSGD and  adjusts sample contribution of each group to mitigate the impact of sample size.

For the group with larger group sample size, we reweight the sample contribution with $\theta^k \propto \frac{1}{\tilde b^k} $ instead of using uniform weight of 1 for all groups, where $\tilde b^k$ is privately estimated (Line \ref{line:privb} in Algorithm \ref{alg:naive}).
Note that $G_B$ in Algorithm \ref{alg:dpsgd} is estimated based on uniform weight of each sample regardless of their group membership.
The sensitivity for  group $k$ is $ C^k={C_0}\times  \theta^k $. The sensitivity for the total population would be $C^0\times\max_k \theta^k$.
The result also matches the idea that we limit the sample contribution of the group with smaller cost of privacy to achieve stronger privacy level w.r.t. the group. However, Na\"ive only considers the group sample size. As we know from previous observation and analysis, the factors that affect the gradient norm and bias due to clipping are more complex than just the group sample size. We will compare with this Na\"ive approach as a baseline in our experiments.

\begin{algorithm}[tb]
	\begin{algorithmic}[1]
        \FOR {$t\in [T]$}
            \STATE Randomly sample a batch of samples $B_t$ with  $|B_t|=b$ from $D$ 
            \FOR {each sample $x_i \in B_t$}
                \STATE $g_i=\triangledown L_i({w}_t)$
            \ENDFOR
            \STATE $\left\{\tilde b^k\right\}_{k \in [K]}=\left\{b^k\right\}_{k \in [K]}+N(0,\sigma_1^2\mathbf I)$
            \label{line:privb}
            \FOR {each group $k \in [K]$}
                \STATE $\theta^k=1\times \frac{b/K}{\tilde b^k }$
            \ENDFOR
            \FOR {each sample $x_i \in B_t$}
                \STATE $\bar{g}_i= \theta^k \times g_i\times\min\left(1,\frac{C_0}{|g_i|}\right)$
            \ENDFOR
            \STATE $C= C_0\times \max\limits_k\theta^k$
            \STATE $\tilde{G}_B=\frac{1}{b}\left(\sum_i \bar{g}_i+N(0,\sigma_2^2C^2\mathbf I)\right)$
            \STATE $\tilde{w}_{t+1}=\tilde{w}_{t}-r \tilde{G}_B$
        \ENDFOR
		\STATE Return ${\tilde{w}_T}$ and accumulated privacy cost $(\epsilon,\delta)$
	\end{algorithmic}	
	\caption{Na\"ive (Dataset $D$, loss function $ \mathcal{L}_D({w})$, learning rate $r$, batch size $b$, noise scales $\sigma_1, \sigma_2$, base clipping bound $C_0$)}
	\label{alg:naive}
\end{algorithm}

\section{Experiments}
\label{sec:eval}
\subsection{Experiment Setup}
\label{sec:setup}
\subsubsection{Datasets}

We use MNIST dataset and replicate the setting in \cite{Bagdasaryan2019}. The original MNIST dataset is a balanced dataset with 60,000 training samples and each class has about 6,000 samples. Class 8 has the most false negatives, hence we  choose it as the artificially underrepresented group (reducing the number of training samples from 5,851 to 500) in the unbalanced MNIST dataset. We compare the underrepresented class 8 with the well-represented class 2 that shares fewest false negatives with the class 8 and therefore can be considered independent. The testing dataset has 10,000 testing samples with about 1,000 for each class.

We also use two census datasets, Adult and Dutch. For both  datasets, we consider ``Sex'' as  the protected attribute and ``Income'' as  decision. For unprotected attributes, we convert  categorical attributes  to  one-hot vectors  and normalize  numerical attributes to $[0,1]$ range. After preprocessing, we have 40 unprotected attributes for Adult and 35 unprotected attributes for Dutch. 
The original Adult dataset  has 45,222 samples (30,527 males and 14,695 females). We sample a balanced Adult dataset with 14,000 males and 14,000 females. The original Dutch dataset is close to balanced with 30,273 males and 30,147 females. We sample an unbalanced Dutch dataset with 30,000 males and 10,000 females. In all settings, we split the census datasets into 80\% training data and 20\% testing data.

\subsubsection{Model}

For the MNIST dataset, we use a neural network with 2 convolutional layers and 2 linear layers with 431K parameters in total. We use learning rate $r=0.01$, batch size $b=256$, and the number of training  epochs is 60.

For the census datasets, we use a logistic regression model with regularization parameter 0.01. We use learning rate $r=1/\sqrt{T}$, batch size $b=256$, and the number of training  epochs is 20. 

\subsubsection{Baseline}
We compare our proposed method DPSGD-F (Algorithm \ref{alg:dpsgd_f}) with the original DPSGD (Algorithm \ref{alg:dpsgd}) and the Na\"ive approach (Algorithm \ref{alg:naive}). 
For each setting, the learning parameters are the same. We set $C_0, \sigma_2$ in DPSGD-F and Na\"ive equal to $C, \sigma$ in DPSGD, respectively. We set $\sigma_1 = 10 \sigma_2$. For the MNIST dataset, we set noise scale $\sigma=0.8$, clipping bound $C=1$, and $\delta=10^{-6}$. 
For the census datasets, we set noise scale $\sigma=1$, clipping bound $C=0.5$, and $\delta=10^{-6}$.
The accumulated privacy budget $\epsilon$ for each setting is computed using the privacy moments accounting method \cite{AbadiCGMMT016}. Because we set $\sigma_1 = 10 \sigma_2$, most of $\epsilon$ is spent on gradients from $\sigma_2$. Only about 0.01 budget is from $\sigma_1$. 
To compare  DPSGD-F and  Na\"ive  with DPSGD under the same privacy budget, the algorithm runs a few less iterations than DPSGD in the last epoch, where the total number of iterations $T= \text{epochs}\times n/b$ in SGD and DPSGD. For DPSGD-F and  Na\"ive, $T$ is 22 and 19 less on the balanced and unbalanced MNIST datasets, respectively; 5 and 11 less on the balanced and unbalanced Adult datasets, respectively; 17 and 9 less on the balanced and unbalanced Dutch datasets, respectively. These differences are very small in proportion to $T$.
All DP models are compared with the non-private SGD when we measure the utility loss due to differential privacy.

\subsubsection{Metric}
We use the test data to measure the model utility and fairness. Based on Definition \ref{def:eoi}, we use reduction in model accuracy for each group between the private SGD and the non-private SGD ($\Delta^k$) as the metric to measure the impact of differential privacy w.r.t. each group. The difference between the impacts on groups ($|\Delta^i-\Delta^j|$ for any $i,j$) measures the level of inequality in utility loss due to differential privacy.  If the impacts for all groups are independent of the protected attribute ($|\Delta^i-\Delta^j| \leq \tau$ for any $i,j$, for example $\tau = 0.05$),  we consider the private SGD has equal reduction in model accuracy w.r.t. each group, i.e. the private SGD achieves equality of impact of differential privacy.  We also report the average loss and average gradient norm
to show the convergence w.r.t. each group during training.

\begin{table*}[]
\small
\centering
\begin{tabular}{|c|c|c|c|c|c|c|c|c|c|c|c|c|}
\hline
                       & \multicolumn{6}{c|}{Average loss}                                                    & \multicolumn{6}{c|}{Average gradient norm}                                           \\ \hline
Dataset                & \multicolumn{2}{c|}{MNIST} & \multicolumn{2}{c|}{Adult} & \multicolumn{2}{c|}{Dutch} & \multicolumn{2}{c|}{MNIST} & \multicolumn{2}{c|}{Adult} & \multicolumn{2}{c|}{Dutch} \\ \hline
Group                  & Class 2      & Class 8     & M            & F           & M            & F           & Class 2      & Class 8     & M            & F           & M            & F           \\ \hline
SGD                    & 0.04         & 0.04        & 0.48         & 0.27        & 0.49         & 0.58        & 0.68         & 4.76        & 0.08         & 0.11        & 0.12         & 0.30        \\ \hline
DPSGD                  & 0.41         & 2.16        & 0.68         & 0.31        & 0.52         & 0.83        & 13.53        & 100.46      & 0.41         & 0.12        & 0.11         & 0.52        \\ \hline
Na\"ive & 3.08         & 1.89        & 0.71         & 0.32        & 0.58         & 0.55        & 0.83         & 0.76        & 0.43         & 0.13        & 0.23         & 0.17        \\ \hline
DPSGD-F            & 0.20         & 0.42        & 0.50         & 0.27        & 0.48         & 0.61        & 1.45         & 2.53        & 0.12         & 0.08        & 0.09         & 0.35        \\ \hline
\end{tabular}
	\caption{The average loss  and  the average  gradient norm w.r.t. groups at the last training epoch on the unbalanced MNIST ($\epsilon=6.55, \delta=10^{-6}$), the unbalanced Adult ($\epsilon=3.1, \delta=10^{-6}$) and  the unbalanced Dutch ($\epsilon=3.29, \delta=10^{-6}$) datasets}
	\label{tbl:posttrain}
\end{table*}

\begin{table*}[tb]
\small
\centering
\begin{tabular}{|c|c|c|c|c|c|c|}
\hline
Dataset                  & \multicolumn{3}{c|}{Balanced}                          & \multicolumn{3}{c|}{Unbalanced}                        \\ \hline
Group                    & Total            & Class 2          & Class 8          & Total            & Class 2          & Class 8          \\ \hline
Sample size              & 60000            & 5958             & 5851             & 54649            & 5958             & 500              \\ \hline
SGD                      & 0.9892           & 0.9932           & 0.9917           & 0.9855           & 0.9903           & 0.9292           \\ \hline
DPSGD vs. SGD            & -0.0494          & -0.0853          & -0.0719          & -0.1081          & -0.0707          & -0.6807          \\ \hline
Na\"ive vs. SGD            & -0.0491          & -0.0891          & -0.0687          & -0.1500          & -0.1512          & -0.1510          \\ \hline
{DPSGD-F vs. SGD} & {-0.0236} & {-0.0339} & {-0.0359} & {-0.0293} & {-0.0281} & {-0.0432} \\ \hline
\end{tabular}
	\caption{Model accuracy w.r.t. class 2 and 8 on the  MNIST dataset (Balanced: $\epsilon=6.23, \delta=10^{-6}$, Unbalanced: $\epsilon=6.55, \delta=10^{-6}$)}
	\label{tbl:mnist}
\end{table*}

\begin{figure*}
     \centering
     \begin{subfigure}[tb]{0.48\textwidth}
         \centering
         \includegraphics[width=\textwidth]{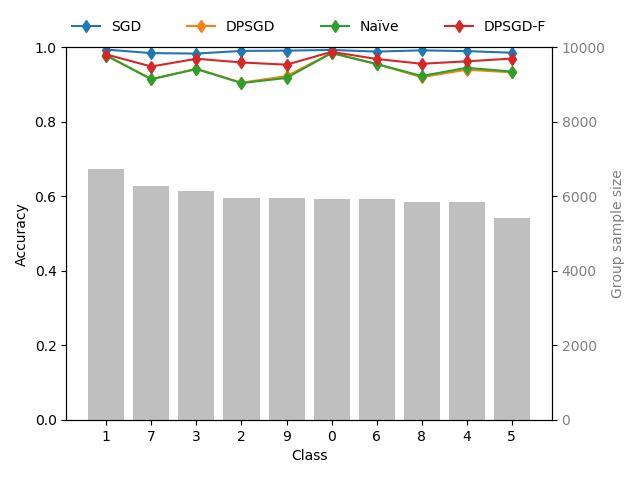}
         \subcaption{Balanced: $\epsilon=6.23, \delta=10^{-6}$}
         \label{fig:mnist_b}
     \end{subfigure}
     \begin{subfigure}[tb]{0.48\textwidth}
         \centering
         \includegraphics[width=\textwidth]{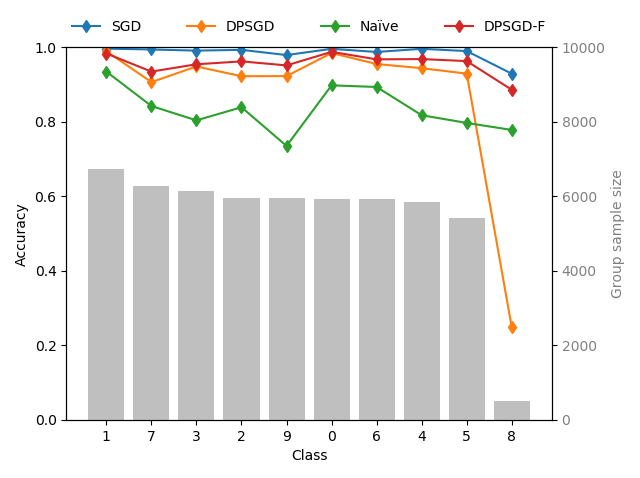}
         \subcaption{Unbalanced: $\epsilon=6.55, \delta=10^{-6}$}
         \label{fig:mnist_ub}
     \end{subfigure}
        \caption{Model accuracy w.r.t. each class  for SGD, DPSGD, Na\"ive and DPSGD-F on the  MNIST dataset}
        \label{fig:mnist}
\end{figure*}

\begin{figure*}
         \centering
         \includegraphics[width=\textwidth]{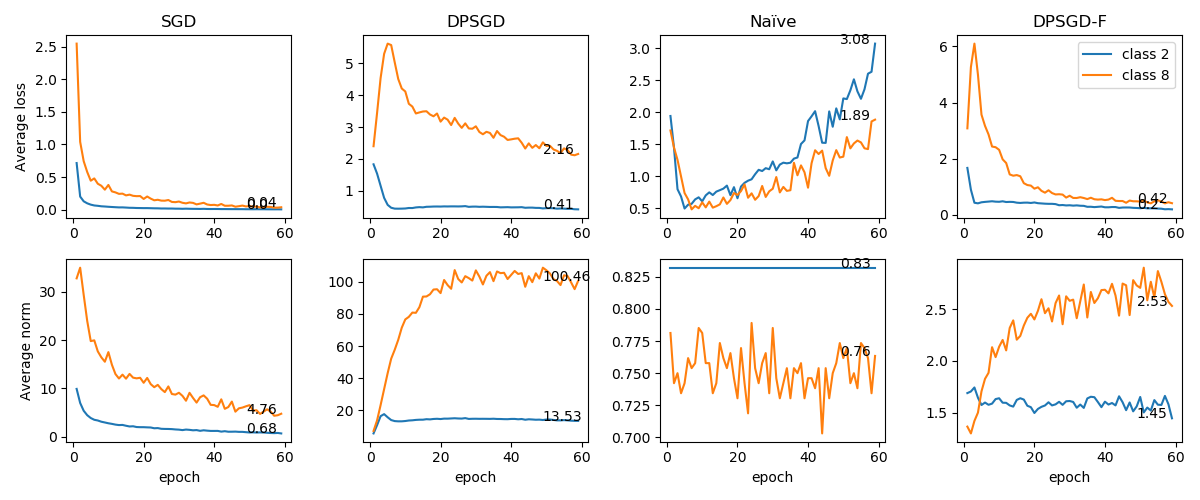}
    	\caption{The average loss  and  the average  gradient norm w.r.t. class 2 and 8 over epochs for SGD, DPSGD, Na\"ive and DPSGD-F on the unbalanced MNIST dataset ($\epsilon=6.55, \delta=10^{-6}$)}
    	\label{fig:group_mnist_post}
\end{figure*}

\begin{figure}
         \centering
         \includegraphics[width=\linewidth]{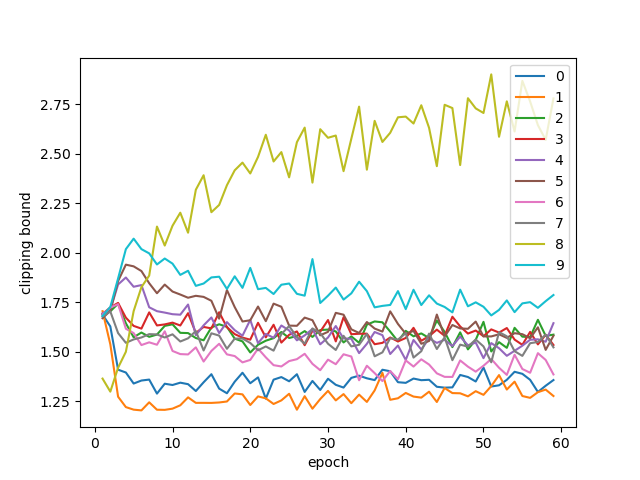}
    	\caption{The clipping bound $C^k$ w.r.t. each class over epochs for DPSGD-F on the unbalanced MNIST dataset (${\epsilon=6.55, \delta=10^{-6}}$)}
    	\label{fig:clipping}
\end{figure}

\begin{table}[tb]
\small
\centering
\begin{tabular}{|c|c|c|c|}
\hline
Group               & Total   & Class 2 & Class 8  \\ \cline{1-4}
Sample size         & 54649   & 5958    & 500      \\ \cline{1-4}
SGD                 & 0.9855  & 0.9903  & 0.9292   \\ \hline
DPSGD ($C=1$) vs. SGD & -0.1081 & -0.0707 & -0.6807  \\ \hline
DPSGD ($C=2$) vs. SGD & -0.0587 & -0.0426 & -0.3286  \\ \hline
DPSGD ($C=3$) vs. SGD & -0.0390 & -0.0232 & -0.2013  \\ \hline
DPSGD ($C=4$) vs. SGD & -0.0286 & -0.0194 & -0.1376  \\ \hline
DPSGD ($C=5$) vs. SGD & -0.0240 & -0.0145 & -0.1099  \\ \hline
DPSGD-F ($C_0=1$) vs. SGD     & -0.0293 & -0.0281 & -0.0432  \\ \hline
\end{tabular}
	\caption{Model accuracy w.r.t. class 2 and 8  for different uniform clipping bound ($C=1,2,3,4,5$) in DPSGD vs. adaptive clipping bound ($C_0=1$) in DPSGD-F  on the unbalanced MNIST dataset (${\epsilon=6.55, \delta=10^{-6}}$)}
	\label{tbl:uni_clip}
\end{table}

\subsection{MNIST Dataset}

Table \ref{tbl:mnist} shows the model accuracy w.r.t. class 2 and 8 on the balanced and unbalanced MNIST datasets.
On the balanced dataset,  each private or non-private model achieves similar accuracy across all groups. 
When we artificially reduce the sample size of class 8, class 8 becomes the minority group in the unbalanced dataset. 
The non-private SGD model  converges to $0.9292$ accuracy on class 8 vs. $0.9903$ accuracy on class 2. The DPSGD model causes $-0.6807$ accuracy loss on class 8 vs. $-0.0707$ on class 2, which exhibits a significant disparate impact on the underrepresented class. 
The Na\"ive approach achieves $-0.1510$ accuracy loss on class 8 vs. $-0.1512$ on class 2, which achieves equal privacy impact. 
Our DPSGD-F algorithm has $-0.0432$ accuracy loss on class 8 vs. $-0.0281$ on class 2, which also achieves equal privacy impact. The total model accuracy also drops less for DPSGD-F ($-0.0293$) than the original DPSGD ($-0.1081$).
Figure \ref{fig:mnist} shows the model accuracy w.r.t. all classes on the MNIST dataset. The difference between DPSGD and DPSGD-F is small and consistent across all classes.

Table \ref{tbl:posttrain} shows the average loss and average gradient norm w.r.t. class 2 and 8 for SGD and different DP models at the last training epoch. 
In DPSGD, the average gradient norm for class 8 is over 100 and  the average loss for class 8 is 2.16. Whereas, in DPSGD-F, the average gradient norm for class 8  is only 2.53 and the average loss for class 8  is only 0.42. The convergence loss and the gradient norm for class 8 are much closer to the ones for class 2 in DPSGD-F.  Figure \ref{fig:group_mnist_post} shows the convergence trend during training. The trend in DPSGD-F is the closest to the trend in SGD among all DP models. It shows our adjusted clipping bound helps to achieve the same group utility loss  regardless of the group sample size.  

Figure \ref{fig:clipping} shows how our adaptive clipping bound changes over epochs in DPSGD-F. Because class 8 has larger clipping bias due to its underrepresented group sample size, {DPSGD-F} gives class 8 a higher clipping bound to increase its sample contribution in the total gradient. The maximal $C^k$ is close to 3. To show that the fair performance of DPSGD-F is not caused by increasing clipping bound uniformly, we run the original DPSGD with increasing clipping bound  from $C=1$ to $C=5$. Table \ref{tbl:uni_clip} shows the level of inequality in utility loss for different clipping bound in DPSGD vs. the adaptive clipping bound in DPSGD-F. Even though increasing clipping bound $C$ in DPSGD can improve the accuracy on class 8, there is still significant difference between the accuracy loss  on class 8 ($-0.1099$ when $C=5$) and the accuracy loss on class 2 ($-0.0145$ when $C=5$). This is because the utility-privacy trade-offs are different for the minority group and the majority group under the same clipping bound. So the inequality in utility loss cannot be removed by simply increasing the clipping bound in DPSGD. On the contrary, {DPSGD-F} achieves equal privacy impact on model accuracy by adjusting the clipping bound for each group according to the utility-privacy trade-off. The group with smaller cost of privacy achieves a stronger level of privacy as a result of adaptive clipping.

\begin{table*}[]
\small
\centering
\begin{tabular}{|c|c|c|c|c|c|c|c|c|c|c|c|c|}
\hline
Dataset         & \multicolumn{3}{c|}{Balanced Adult}                    & \multicolumn{3}{c|}{Unbalanced Adult}                  & \multicolumn{3}{c|}{Balanced Dutch}                    & \multicolumn{3}{c|}{Unbalanced Dutch}                 \\ \hline
Group           & Total                & M                & F                & Total                 & M               & F                &   Total               & M                & F                &  Total                & M               & F                \\ \hline
Sample size     & 28000            & 14000            & 14000            & 45222            & 30527           & 14695            & 60420            & 30273            & 30147            & 40000            & 30000           & 10000            \\ \hline
SGD             & 0.824           & 0.748           & 0.899           & 0.809           & 0.761           & 0.911           & 0.787           & 0.801           & 0.774           & 0.802           & 0.834          & 0.706           \\ \hline
DPSGD vs. SGD    & -0.036          & -0.054          & -0.019          & -0.059          & -0.074          & -0.028          & -0.100          & -0.153          & -0.046          & -0.124          & -0.086          & -0.240            \\ \hline
Na\"ive vs. SGD    & -0.036          & -0.054          & -0.019          & -0.059          & -0.074         & -0.028          & -0.100          & -0.155          & -0.046          & -0.101           & -0.142         & 0.025           \\ \hline
DPSGD-F vs. SGD  & {-0.009} & {-0.014} & {-0.005} & {-0.025} & {-0.029} & {-0.016} & {-0.013} & {-0.016} & {-0.009} & {-0.032} & {-0.028} & {-0.044} \\ \hline
\end{tabular}
	\caption{Model accuracy w.r.t. the total population and each group on the Adult  and Dutch  datasets (Balanced Adult (sampled): $\epsilon=3.99, \delta=10^{-6}$, Unbalanced Adult (original): $\epsilon=3.1, \delta=10^{-6}$, Balanced Dutch (original): $\epsilon=2.66, \delta=10^{-6}$, Unbalanced Dutch (sampled): $\epsilon=3.29, \delta=10^{-6}$)}
	\label{tbl:adult}
\end{table*}

\begin{figure*}
         \centering
         \includegraphics[width=\textwidth]{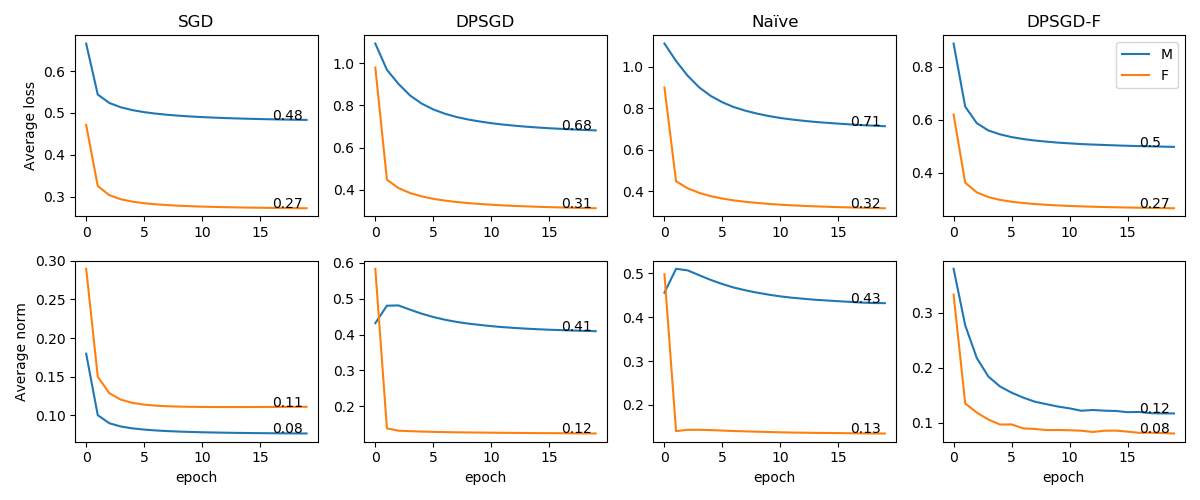}
    	\caption{The average loss  and  the average  gradient norm w.r.t. each group over epochs for SGD, DPSGD, Na\"ive and DPSGD-F on the unbalanced Adult dataset ($\epsilon=3.1, \delta=10^{-6}$)}
    	\label{fig:group_adult_post}
\end{figure*}

\begin{figure*}
         \centering
         \includegraphics[width=\textwidth]{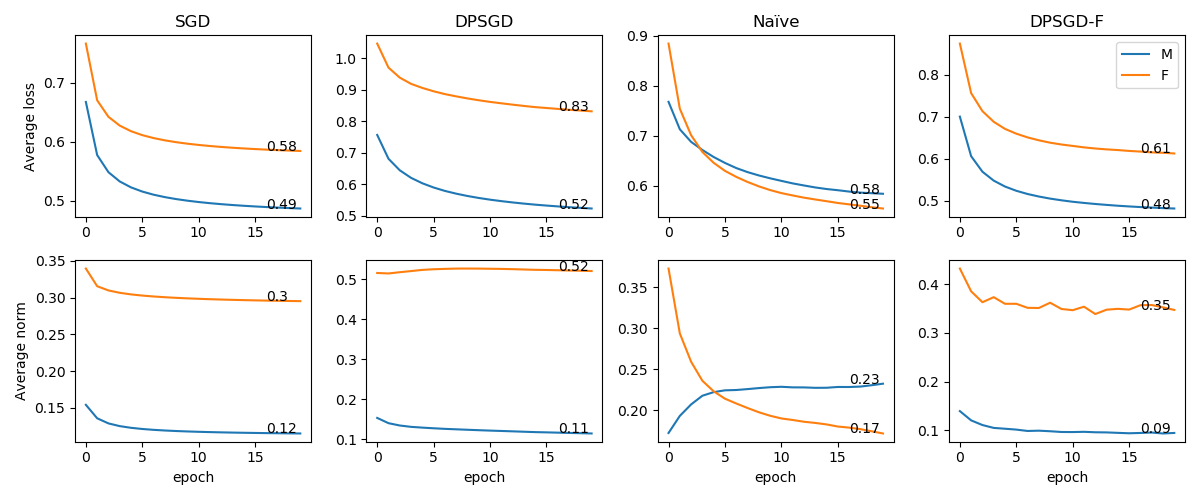}
    	\caption{The average loss  and  the average  gradient norm w.r.t. each group over epochs for SGD, DPSGD, Na\"ive and DPSGD-F on the unbalanced Dutch dataset ($\epsilon=3.29, \delta=10^{-6}$)}
    	\label{fig:group_dutch_post}
\end{figure*}

\subsection{Adult and Dutch Datasets}

Table \ref{tbl:adult} shows the model accuracy w.r.t. male and female on the balanced and unbalanced Adult and Dutch datasets. 
The clipping biases for both census datasets are not primarily decided by group sample size. We observe disparate impact on DPSGD in comparison to SGD against the male group, even though the male group is not underrepresented. 
The Na\"ive approach does not work at all to achieve equal privacy impact in this case, as the importance of group sample size is not as much as in the MNIST dataset. 
There are still other factors that affect the  gradient norm and the clipping bias w.r.t. each group.
DPSGD-F can achieve similar accuracy loss for male and female in all four settings. It shows the effectiveness of our approach.

Table \ref{tbl:posttrain} shows the average loss and average gradient norm w.r.t. male and female for SGD and different DP models at the last training epoch.
On the unbalanced Adult dataset, the average gradient norm in DPSGD for male  is 5 times of the one in SGD and  the average loss in DPSGD for male  is 50\% more than the one in SGD. Whereas, in DPSGD-F, the average gradient norm and  the average loss for the male group  are much closer to the ones in SGD.  
Similar to the Adult dataset, on the  unbalanced Dutch dataset, the average gradient norm and the average loss in DPSGD-F for the male group  are much closer to the ones in SGD. 
Figure \ref{fig:group_adult_post} and \ref{fig:group_dutch_post} show the convergence trends  on the unbalanced Adult and Dutch datasets during training. 
The trends in DPSGD-F is the closest to the trends in SGD among all DP models.
It shows that our adjusted clipping bound helps to achieve the same group utility loss.

\section{Conclusion and Future Work}

Gradient clipping and random noise addition, which are the core techniques in differentially private
SGD, disproportionately affect underrepresented and complex classes and subgroups. As a
consequence, DPSGD has disparate impact: the accuracy of a model trained using
DPSGD tends to decrease more on these classes and subgroups vs. the original, non-private model.
If the original model is unfair in the sense that its accuracy is not the same across all subgroups,
DPSGD exacerbates this unfairness.
In this work, we propose DPSGD-F to remove the potential disparate impact of differential privacy on the protected group. 
DPSGD-F adjusts the contribution of samples in a group depending on the group clipping bias such that differential privacy has no disparate impact on group utility.
Our experimental evaluation shows how group sample size and group clipping bias affect the impact of differential privacy in DPSGD, and how adaptive  clipping for each group helps to mitigate the disparate impact caused by differential privacy in DPSGD-F.
Gradient clipping in the non-private context may improve the model robustness against outliers. However, examples in the minority group are not outliers. They should not be ignored by the (private) learning model.
In future work, we can further improve our adaptive clipping method from group-wise adaptive clipping to element-wise (from user and/or parameter perspectives) adaptive clipping, so the model can be fair even to the unseen minority class.

\section*{Acknowledgments}
This work was supported in part by NSF 1502273, 1920920, 1937010 and 1946391.

\bibliographystyle{named}

\begin{thebibliography}{}
	
	\bibitem[\protect\citeauthoryear{Abadi \bgroup \em et al.\egroup
	}{2016}]{AbadiCGMMT016}
	Martin Abadi, Andy Chu, Ian~J. Goodfellow, H.~Brendan McMahan, Ilya Mironov,
	Kunal Talwar, and Li~Zhang.
	\newblock Deep learning with differential privacy.
	\newblock In {\em Proceedings of CCS, Vienna, Austria, October 24-28, 2016}, pages
	308--318, 2016.
	
	\bibitem[\protect\citeauthoryear{Amin \bgroup \em et al.\egroup
	}{2019}]{AminKMV19}
	Kareem Amin, Alex Kulesza, Andres~Mu{\~{n}}oz Medina, and Sergei Vassilvitskii.
	\newblock Bounding user contributions: {A} bias-variance trade-off in
	differential privacy.
	\newblock In {\em Proceedings of the 36th International Conference on Machine
		Learning, {ICML}, 9-15 June 2019, Long Beach, California, {USA}}, pages
	263--271, 2019.
	
	\bibitem[\protect\citeauthoryear{Bagdasaryan \bgroup \em et al.\egroup
	}{2019}]{Bagdasaryan2019}
	Eugene Bagdasaryan, Omid Poursaeed, and Vitaly Shmatikov.
	\newblock Differential privacy has disparate impact on model accuracy.
	\newblock In  {\em  NeurIPS 2019, 8-14 December 2019,
		Vancouver, BC, Canada}, pages 15453--15462, 2019.
	
	\bibitem[\protect\citeauthoryear{Bassily \bgroup \em et al.\egroup
	}{2018}]{DBLP:conf/nips/BassilyTT18}
	Raef Bassily, Abhradeep~Guha Thakurta, and Om~Dipakbhai Thakkar.
	\newblock Model-agnostic private learning.
	\newblock In {\em  NeurIPS 2018, 3-8
		December 2018, Montr{\'{e}}al, Canada}, pages 7102--7112, 2018.
	
	\bibitem[\protect\citeauthoryear{Beutel \bgroup \em et al.\egroup
	}{2017}]{Beutel2017Data}
	Alex Beutel, Jilin Chen, Zhe Zhao, and Ed~H. Chi.
	\newblock Data decisions and theoretical implications when adversarially
	learning fair representations.
	\newblock In {\em FAT/ML}, 2017.
	
	\bibitem[\protect\citeauthoryear{Calders \bgroup \em et al.\egroup
	}{2009}]{Calders2009Building}
	T.~Calders, F.~Kamiran, and M.~Pechenizkiy.
	\newblock Building classifiers with independency constraints.
	\newblock In {\em 2009 IEEE International Conference on Data Mining Workshops},
	pages 13--18, 2009.
	
	\bibitem[\protect\citeauthoryear{Chaudhuri \bgroup \em et al.\egroup
	}{2011}]{DBLP:journals/jmlr/ChaudhuriMS11}
	Kamalika Chaudhuri, Claire Monteleoni, and Anand~D. Sarwate.
	\newblock Differentially private empirical risk minimization.
	\newblock {\em J. Mach. Learn. Res.}, 12:1069--1109, 2011.
	
	\bibitem[\protect\citeauthoryear{Cummings \bgroup \em et al.\egroup
	}{2019}]{CummingsGKM19}
	Rachel Cummings, Varun Gupta, Dhamma Kimpara, and Jamie Morgenstern.
	\newblock On the compatibility of privacy and fairness.
	\newblock In {\em Adjunct Publication of the 27th Conference on User Modeling,
		Adaptation and Personalization, {UMAP} 2019, Larnaca, Cyprus, June 09-12,
		2019}, pages 309--315, 2019.
	
	\bibitem[\protect\citeauthoryear{Ding \bgroup \em et al.\egroup
	}{2020}]{ding2020differentially}
	Jiahao Ding, Xinyue Zhang, Xiaohuan Li, Junyi Wang, Rong Yu, and Miao Pan.
	\newblock Differentially private and fair classification via calibrated
	functional mechanism.
	\newblock In {\em AAAI}, 2020.
	
	\bibitem[\protect\citeauthoryear{Du \bgroup \em et al.\egroup }{2019}]{Du2019}
	Min Du, Ruoxi Jia, and Dawn Song.
	\newblock Robust anomaly detection and backdoor attack detection via
	differential privacy.
	\newblock {\em CoRR}, abs/1911.07116, 2019.
	
	\bibitem[\protect\citeauthoryear{Duchi \bgroup \em et al.\egroup
	}{2013}]{DBLP:conf/focs/DuchiJW13}
	John~C. Duchi, Michael~I. Jordan, and Martin~J. Wainwright.
	\newblock Local privacy and statistical minimax rates.
	\newblock In {\em 54th Annual {IEEE} Symposium on Foundations of Computer
		Science, {FOCS} 2013, 26-29 October, 2013, Berkeley, CA, {USA}}, pages
	429--438, 2013.
	
	\bibitem[\protect\citeauthoryear{Dwork \bgroup \em et al.\egroup
	}{2006}]{Dwork2006}
	Cynthia Dwork, Frank McSherry, Kobbi Nissim, and Adam~D. Smith.
	\newblock Calibrating noise to sensitivity in private data analysis.
	\newblock In {\em Theory of Cryptography, Third}, pages 265--284, 2006.
	
	\bibitem[\protect\citeauthoryear{Dwork \bgroup \em et al.\egroup
	}{2012}]{DBLP:conf/innovations/DworkHPRZ12}
	Cynthia Dwork, Moritz Hardt, Toniann Pitassi, Omer Reingold, and Richard~S.
	Zemel.
	\newblock Fairness through awareness.
	\newblock In {\em Innovations in Theoretical Computer Science}, pages 214--226,
	2012.
	
	\bibitem[\protect\citeauthoryear{Edwards and
		Storkey}{2016}]{Edwards2015Censoring}
	Harrison Edwards and Amos~J. Storkey.
	\newblock Censoring representations with an adversary.
	\newblock In  {\em 4th International
		Conference on Learning Representations, {ICLR} 2016, San Juan, Puerto Rico,
		May 2-4, 2016, Conference Track Proceedings}, 2016.
	
	\bibitem[\protect\citeauthoryear{Ekstrand \bgroup \em et al.\egroup
	}{2018}]{DBLP:conf/fat/EkstrandJM18}
	Michael~D. Ekstrand, Rezvan Joshaghani, and Hoda Mehrpouyan.
	\newblock Privacy for all: Ensuring fair and equitable privacy protections.
	\newblock In {\em Conference on Fairness, Accountability and Transparency},
	pages 35--47, 2018.
	
	\bibitem[\protect\citeauthoryear{Feldman \bgroup \em et al.\egroup
	}{2015}]{feldman2015}
	Michael Feldman, Sorelle~A. Friedler, John Moeller, Carlos Scheidegger, and
	Suresh Venkatasubramanian.
	\newblock Certifying and {{Removing Disparate Impact}}.
	\newblock In {\em Proceedings of the 21th {{ACM SIGKDD International
				Conference}} on {{Knowledge Discovery}} and {{Data Mining}} - {{KDD}} '15},
	New York, NY, USA, 2015.
	
	\bibitem[\protect\citeauthoryear{Hajian \bgroup \em et al.\egroup
	}{2015}]{DBLP:journals/datamine/HajianDMPG15}
	Sara Hajian, Josep Domingo{-}Ferrer, Anna Monreale, Dino Pedreschi, and Fosca
	Giannotti.
	\newblock Discrimination- and privacy-aware patterns.
	\newblock {\em Data Min. Knowl. Discov.}, 29(6):1733--1782, 2015.
	
	\bibitem[\protect\citeauthoryear{Hardt \bgroup \em et al.\egroup
	}{2016}]{Hardt2016Equality}
	Moritz Hardt, Eric Price, and Nathan Srebro.
	\newblock Equality of opportunity in supervised learning.
	\newblock In {\em NeurIPS}, 2016.
	
	\bibitem[\protect\citeauthoryear{Jagielski \bgroup \em et al.\egroup
	}{2019}]{JagielskiKMORSU19}
	Matthew Jagielski, Michael~J. Kearns, Jieming Mao, Alina Oprea, Aaron Roth,
	Saeed Sharifi{-}Malvajerdi, and Jonathan Ullman.
	\newblock Differentially private fair learning.
	\newblock In {\em Proceedings of the 36th International Conference on Machine
		Learning, {ICML} 2019, 9-15 June 2019, Long Beach, California, {USA}}, pages
	3000--3008, 2019.
	
	\bibitem[\protect\citeauthoryear{Jaiswal and Provost}{2019}]{Jaiswal2019}
	Mimansa Jaiswal and Emily~Mower Provost.
	\newblock Privacy enhanced multimodal neural representations for emotion
	recognition.
	\newblock {\em CoRR}, abs/1910.13212, 2019.
	
	\bibitem[\protect\citeauthoryear{Kamiran and Calders}{2009}]{kamiran2009}
	Faisal Kamiran and Toon Calders.
	\newblock Classifying without discriminating.
	\newblock In {\em 2009 2nd {{International Conference}} on {{Computer}},
		{{Control}} and {{Communication}}}, pages 1--6. {IEEE}, February 2009.
	
	\bibitem[\protect\citeauthoryear{Kamiran and Calders}{2011}]{Kamiran2012Data}
	Faisal Kamiran and Toon Calders.
	\newblock Data preprocessing techniques for classification without
	discrimination.
	\newblock {\em Knowl. Inf. Syst.}, 33(1):1--33, 2011.
	
	\bibitem[\protect\citeauthoryear{Kamiran \bgroup \em et al.\egroup
	}{2010}]{Kamiran2010Discrimination}
	F.~Kamiran, T.~Calders, and M.~Pechenizkiy.
	\newblock Discrimination aware decision tree learning.
	\newblock In {\em 2010 IEEE International Conference on Data Mining}, pages
	869--874, 2010.
	
	\bibitem[\protect\citeauthoryear{Kamishima \bgroup \em et al.\egroup
	}{2011}]{Kamishima2011FairnessAware}
	T.~Kamishima, S.~Akaho, and J.~Sakuma.
	\newblock Fairness-aware learning through regularization approach.
	\newblock In {\em 2011 IEEE 11th International Conference on Data Mining
		Workshops}, pages 643--650, 2011.
	
	\bibitem[\protect\citeauthoryear{Kearns \bgroup \em et al.\egroup
	}{2018}]{DBLP:conf/icml/KearnsNRW18}
	Michael~J. Kearns, Seth Neel, Aaron Roth, and Zhiwei~Steven Wu.
	\newblock Preventing fairness gerrymandering: Auditing and learning for
	subgroup fairness.
	\newblock In {\em Proceedings of the 35th International Conference on Machine
		Learning, {ICML} 2018, Stockholmsm{\"{a}}ssan, Stockholm, Sweden, July 10-15,
		2018}, pages 2569--2577, 2018.
	
	\bibitem[\protect\citeauthoryear{Krasanakis \bgroup \em et al.\egroup
	}{2018}]{KrasanakisXPK18}
	Emmanouil Krasanakis, Eleftherios~Spyromitros Xioufis, Symeon Papadopoulos, and
	Yiannis Kompatsiaris.
	\newblock Adaptive sensitive reweighting to mitigate bias in fairness-aware
	classification.
	\newblock In {\em Proceedings of the 2018 World Wide Web Conference on World
		Wide Web, {WWW} 2018, Lyon, France, April 23-27, 2018}, pages 853--862, 2018.
	
	\bibitem[\protect\citeauthoryear{Lee and Kifer}{2018}]{LeeK18}
	Jaewoo Lee and Daniel Kifer.
	\newblock Concentrated differentially private gradient descent with adaptive
	per-iteration privacy budget.
	\newblock In {\em Proceedings of the 24th {ACM} {SIGKDD} International
		Conference on Knowledge Discovery {\&} Data Mining, {KDD} 2018, London, UK,
		August 19-23, 2018}, pages 1656--1665, 2018.
	
	\bibitem[\protect\citeauthoryear{Madras \bgroup \em et al.\egroup
	}{2018}]{Madras2018Learning}
	David Madras, Elliot Creager, Toniann Pitassi, and Richard~S. Zemel.
	\newblock Learning adversarially fair and transferable representations.
	\newblock In {\em Proceedings of
		the 35th International Conference on Machine Learning, {ICML} 2018,
		Stockholmsm{\"{a}}ssan, Stockholm, Sweden, July 10-15, 2018}, volume~80 of
	{\em Proceedings of Machine Learning Research}, pages 3381--3390. {PMLR},
	2018.
	
	\bibitem[\protect\citeauthoryear{Phan \bgroup \em et al.\egroup
	}{2017}]{PhanWHD17}
	NhatHai Phan, Xintao Wu, Han Hu, and Dejing Dou.
	\newblock Adaptive laplace mechanism: Differential privacy preservation in deep
	learning.
	\newblock In {\em 2017 {IEEE} International Conference on Data Mining, {ICDM}
		2017, New Orleans, LA, USA, November 18-21, 2017}, pages 385--394, 2017.
	
	\bibitem[\protect\citeauthoryear{Pichapati \bgroup \em et al.\egroup
	}{2019}]{AdaCliP}
	Venkatadheeraj Pichapati, Ananda~Theertha Suresh, Felix~X. Yu, Sashank~J.
	Reddi, and Sanjiv Kumar.
	\newblock Adaclip: Adaptive clipping for private {SGD}.
	\newblock {\em CoRR}, abs/1908.07643, 2019.
	
	\bibitem[\protect\citeauthoryear{Thakkar \bgroup \em et al.\egroup
	}{2019}]{Thakkar2019}
	Om~Thakkar, Galen Andrew, and H.~Brendan McMahan.
	\newblock Differentially private learning with adaptive clipping.
	\newblock {\em CoRR}, abs/1905.03871, 2019.
	
	\bibitem[\protect\citeauthoryear{Xie \bgroup \em et al.\egroup
	}{2017}]{xie2017}
	Qizhe Xie, Zihang Dai, Yulun Du, Eduard Hovy, and Graham Neubig.
	\newblock Controllable {{Invariance}} through {{Adversarial Feature Learning}}.
	\newblock {\em Advances in Neural Information Processing Systems 30 (NIPS
		2017)}, 2017.
	
	\bibitem[\protect\citeauthoryear{Xu \bgroup \em et al.\egroup
	}{2019}]{WWWXuYW19}
	Depeng Xu, Shuhan Yuan, and Xintao Wu.
	\newblock Achieving differential privacy and fairness in logistic regression.
	\newblock In {\em Companion of The 2019 World Wide Web Conference, {WWW} 2019,
		San Francisco, CA, USA, May 13-17, 2019}, pages 594--599, 2019.
	
	\bibitem[\protect\citeauthoryear{Zafar \bgroup \em et al.\egroup
	}{2017}]{Zafar2017Fairness}
	Muhammad~Bilal Zafar, Isabel Valera, Manuel~Gomez Rodriguez, and Krishna~P.
	Gummadi.
	\newblock Fairness constraints: Mechanisms for fair classification.
	\newblock In {\em AISTATS}, 2017.
	
	\bibitem[\protect\citeauthoryear{Zhang \bgroup \em et al.\egroup
	}{2017}]{Zhang2017Causal}
	Lu~Zhang, Yongkai Wu, and Xintao Wu.
	\newblock A causal framework for discovering and removing direct and indirect
	discrimination.
	\newblock IJCAI'17, pages 3929--3935, 2017.
	
	\bibitem[\protect\citeauthoryear{Zhang \bgroup \em et al.\egroup
	}{2018}]{Zhang2018Mitigating}
	Brian~Hu Zhang, Blake Lemoine, and Margaret Mitchell.
	\newblock Mitigating unwanted biases with adversarial learning.
	\newblock In {\em AAAI Conference on AI, Ethics and Society}, 2018.
	
\end{thebibliography}

\end{document}